\pdfoutput=1

\documentclass[11pt]{article}

\usepackage[]{acl}

\usepackage{times}
\usepackage{latexsym}

\usepackage[T1]{fontenc}
\usepackage{amsfonts}

\usepackage[utf8]{inputenc}

\usepackage{microtype}

\usepackage{amsmath}
\usepackage{graphicx}

\usepackage{float}
\usepackage{pgfplots}
\usepackage{subfigure}
\usepackage{booktabs}
\usepackage{multicol}
\usepackage{multirow}

\title{Sparse Fuzzy Attention for Structured Sentiment Analysis}

\author{Letian Peng$^{1,2,3,\dag}$, Zuchao Li$^{1,2,3,\dag}$, and Hai Zhao$^{1,2,3}$\thanks{$\ $  Corresponding author. $^\dag$ These authors made equal contribution. This work was supported by Key Projects of National Natural Science Foundation of China (U1836222 and
61733011).}\\

$^{1}$Department of Computer Science and Engineering, Shanghai Jiao Tong University \\
	$^{2}$Key Laboratory of Shanghai Education Commission for Intelligent Interaction \\ and Cognitive Engineering, Shanghai Jiao Tong University, Shanghai, China\\
	$^{3}$MoE Key Lab of Artificial Intelligence, AI Institute, Shanghai Jiao Tong University \\
  {\tt \small \{zxc-00,charlee\}@sjtu.edu.cn, zhaohai@cs.sjtu.edu.cn}}

\begin{document}
\maketitle
\begin{abstract}
Attention scorers have achieved success in parsing tasks like semantic and syntactic dependency parsing. However, in tasks modeled into parsing, like structured sentiment analysis, "dependency edges" are very sparse which hinders parser performance. Thus we propose a sparse and fuzzy attention scorer with pooling layers which improves parser performance and sets the new state-of-the-art on structured sentiment analysis. We further explore the parsing modeling on structured sentiment analysis with second-order parsing and introduce a novel sparse second-order edge building procedure that leads to significant improvement in parsing performance.
\end{abstract}

\section{Introduction}

\begin{figure}
    \centering
    \includegraphics[width=0.5\textwidth]{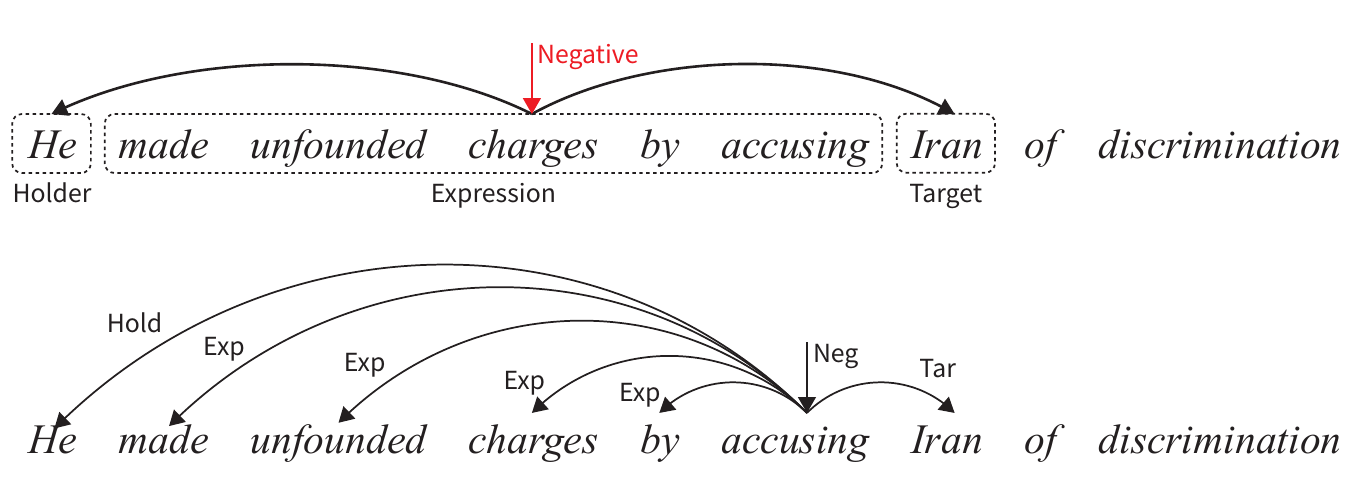}
    \caption{Structured sentiment analysis example and its dependency parsing formulation.}
    \label{fig:ex}
\end{figure}

Structured Sentiment Analysis (SSA) can be formulated into tuple extraction from the context in natural language processing. As shown in Figure~\ref{fig:ex}, a tuple $(p, e, h, t)$ can represent the structure of sentiment in context. Polarity $p$ is expressed by expression $e$ from holder $h$ towards target $t$.

The conventional solution for structured sentiment analysis \citep{choi-etal-2006-joint,yang-cardie-2012-extracting,katiyar-cardie-2016-investigating,DBLP:journals/is/ZhangWF19} breaks down the construction of the entire tuple into span extraction, relationship extraction, and labeling. A unified scenario is recently proposed for structured sentiment analysis and leads to much improvement in model performance. \citet{DBLP:conf/acl/BarnesKOOV20} model structured sentiment analysis as dependency parsing and applies a parser from dependency parsing for the formulated task as in Figure~\ref{fig:ex}. More related works can be referred to Appendix~\ref{app:rel}.

Though the current dependency parser enables the unification of span and relationship prediction procedure, which benefits model performance, we should notice that conventional parsers might not work as efficient as in real dependency parsing when parsing graphs that are formulated into dependency parsing. A key point is that edges in dependency parsing are much denser than in structured sentiment analysis as parsing. For dependency parsing, each component in context is a dependent of a dependency edge. However, this is not the case for structured sentiment analysis as parsing since only components in a sentiment tuple are dependents of edges.

Understanding the sparsity of dependency edges, we propose a sparse attention scorer which concentrates on partials selected by the model for edge scoring. Also, attention for dependents in structured sentiment analysis as parsing is of continuous nature as dependencies are annotated for all components in expressions, holders and targets. Thus, we propose a novel Sparse Fuzzy Attention (SFA) to independently select heads and dependents with sparse and continuous distribution. SFA can be further applied for high-order scorers to further improve model performance as sparsity and continuity also occur in high-order relations.

Experiments on datasets of multiple languages show our SFA is capable to improve parser performance significantly on structured sentiment analysis as parsing. SFA leads to at most $8.0$ and on average $3.2$ F1 score improvement on sentiment graph parsing. For structured sentiment analysis metrics, SFA improves the current state-of-the-art model by $3.7$ for span (holder, target, and expression) F1 score and $2.1$ for targeted F1 score on average.

\section{Sparse Fuzzy Attention}

\subsection{Background}

Biaffine \cite{DBLP:conf/iclr/DozatM17} is a scorer widely used in parsing tasks and other tasks formulated as parsing. Also, biaffine takes the role for edge scoring in the previous state-of-the-art models. Given representations for components in context as heads and dependents $H^h$, $H^d$, biaffine scorer calculates scores for edges and labels with two weight tensors $W^{e}\in \mathbb{R}^{d\times d}$, $W^{l}\in \mathbb{R}^{d\times c \times d}$.
\begin{equation*}
    \centering
    \begin{aligned}
    S^e_{ij}, S^l_{ij} &= (H_i^h)^\mathrm{T}W^eH_j^d, (H_i^h)^\mathrm{T}W^lH_j^d
    \end{aligned}
\end{equation*}
where $S^e \in \mathbb{R}^{n\times n}$ represents edge existence probability and $S^l \in \mathbb{R}^{n\times n \times c}$ represents label type probability. Here $d$ refers to hidden size of representation, $c$ refers to number of label classes, $n$ refers to length of context and $S_{ij}$ consists of scores on edges from $i$-th component to $j$-th component.

\subsection{Encoding}

For a context $C$ of $n$ components $[w_1, w_2, \cdots, w_n]$, we first embed them into tensors with their features including POS, character, and lemma and then feed these tensors into multiple layers of bidirectional long short term memory network (BiLSTM) \cite{DBLP:journals/neco/HochreiterS97}.

\begin{equation*}
    \centering
    \begin{aligned}
    H &= \textrm{Embed}(C) \\
    H &:= \textrm{BiLSTM}(H)
    \end{aligned}
\end{equation*}

Finally, two linear layers are used to project contextualized representations to head and dependent representations.

\begin{equation*}
    \centering
    \begin{aligned}
    H^h, H^d &= W^hH + b^h, W^dH + b^d
    \end{aligned}
\end{equation*}
where weight matrices $W^h, W^d \in \mathbb{R}^{d\times d}$ and biases $b^h, b^d \in \mathbb{R}^d$.

\subsection{Attention Scoring}

\begin{figure}
    \centering
    \includegraphics[width=0.35\textwidth]{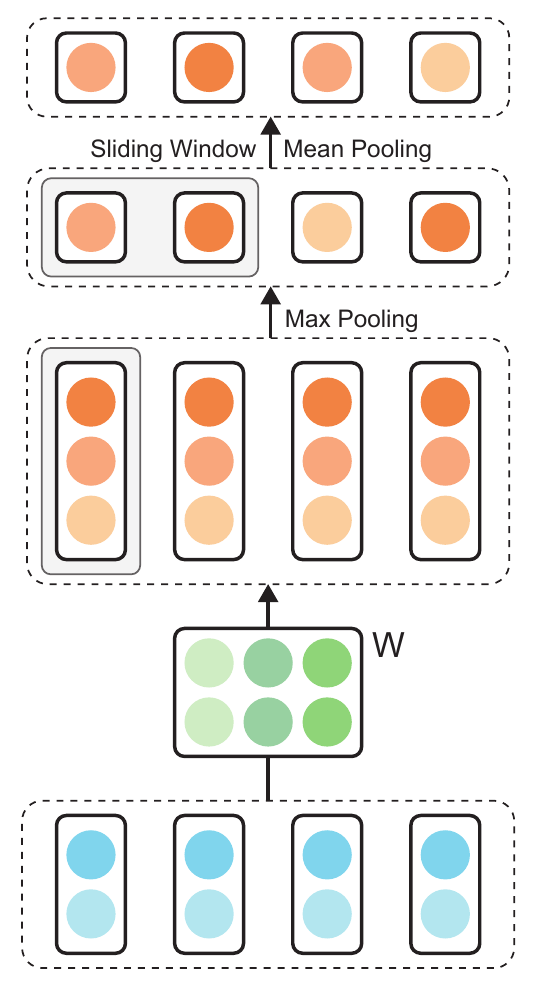}
    \caption{The sparse fuzzy attention}
    \label{fig:sfa}
\end{figure}

We first use the dot product to produce base attention scores. To do this, we project head and dependent representations into different class representations.

\begin{equation*}
    \centering
    \begin{aligned}
    H^{p_i;h}, H^{p_i;d} &= W^{p_i}H^h + b^{p_i}, W^{p_i}H^d + b^{p_i}
    \end{aligned}
\end{equation*}
where $H^{p_i;h} \in \mathbb{R}^{n\times d}$ and $H^{p_i;d} \in \mathbb{R}^{n\times d}$ represent head and dependency representations respectively on $i$-th class. Thus, there exists $c$ matrices and biases for representation projection. Then dot product is used to get scores on edges for each class and these scores are concatenated class-wise together.

\begin{equation*}
    \centering
    \begin{aligned}
    S^{p_i} &= H^{p_i;h}(H^{p_i;d})^\mathrm{T}\\
    S &= [S^{p_1}||S^{p_2}||\cdots||S^{p_c}]
    \end{aligned}
\end{equation*}

Here $c$ is $1$ for edge scorer and the number of label classes for label scorer.

We then describe how we get the fuzzy attention scores for sparse heads and dependents. For head and dependent representations for $i$-th class, we use $W^{p_i;h}, W^{p_i;d} \in \mathbb{R}^{d\times a}$ and $b^{p_i;h}, b^{p_i;d} \in \mathbb{R}^{a}$ to project them into attention scores on multiple heads. 

\begin{equation*}
    \centering
    \begin{aligned}
    E^{p_i;h} &= W^{p_i;h}H^{p_i;h}+b^{p_i;h}\\
    E^{p_i;d} &= W^{p_i;d}H^{p_i;d}+b^{p_i;d}\\
    A_j^{p_i;h} &= \frac{\textrm{exp}(E_j^{p_i;h})}{\sum^n_{k=1}\textrm{exp}(E_k^{p_i;h})}\\
    A_j^{p_i;d} &= \frac{\textrm{exp}(E_j^{p_i;d})}{\sum^n_{k=1}\textrm{exp}(E_k^{p_i;d})}
    \end{aligned}
\end{equation*}

Here $a$ refers to the number of attention heads. Attention scores are then accumulated by max pooling to let the model attend to components selected by different attention heads.

\begin{equation*}
    \centering
    \begin{aligned}
    A_j^{p_i;h} &:= \textrm{max}(A_{j1}^{p_i;h}, A_{j2}^{p_i;h}, \cdots, A_{ja}^{p_i;h})\\
    A_j^{p_i;d} &:= \textrm{max}(A_{j1}^{p_i;d}, A_{j2}^{p_i;d}, \cdots, A_{ja}^{p_i;d})
    \end{aligned}
\end{equation*}

We then use sequence-level mean pooling (by sliding widow) for attention scores to make attention scores fuzzy and thus better concentrate on multiple series of heads or dependents.

\begin{equation*}
    \centering
    \begin{aligned}
    A_j^{p_i;h}, A_j^{p_i;d} &:= \frac{1}{t} \sum^{t-1}_{k=0} A_{j+k}^{p_i;h}, \frac{1}{t} \sum^{t-1}_{k=0} A_{j+k}^{p_i;d}
    \end{aligned}
\end{equation*}

Final, attention for dependencies of $i$-th class is calculated by element-wise attention score multiple. Class-level attention scores are concatenated together for entire attention scores.

\begin{equation*}
    \centering
    \begin{aligned}
    A_{jk}^{p_i} &= A_j^{p_i;h}*A_k^{p_i;d}\\
    A &= [A^{p_1}||A^{p_2}||\cdots||A^{p_c}]\\
    S &:= S * A
    \end{aligned}
\end{equation*}

\subsection{High Order Attention Scoring}

Scoring high-order relation is another efficient mechanism for dealing with sparse distribution of edges. We follow the method of \cite{DBLP:conf/acl/WangHT19} to apply mean field variation inference to induct edges with high order relations.

\begin{equation*}
    \centering
    \begin{aligned}
    S^e := MFVI(S^e, V^{sib}, V^{cop}, V^{grp})
    \end{aligned}
\end{equation*}
where $V^{sib}, V^{cop}, V^{grp} \in \mathbb{R}^{n\times n \times n}$ refer to scores for existence probability of second-order relations sibling, co-parent and grandparent. The specific calculating process is rather complex for elaborate description and can be referred to \cite{DBLP:conf/acl/WangHT19}. This inference is repeated multiple times to get final edge scores.

Conventional second-order scorer uses a matrix $W\in \mathbb{R}^{d\times d \times d}$ for scoring $V$, we make second-order attention score sparser and more continuous with our SFA mechanism.

\begin{equation*}
    \centering
    \begin{aligned}
    V &= H^h(H^m)^\mathrm{T}WH^d \\
    A^h, A^m, A^d &= \textrm{SFA}(H^h), \textrm{SFA}(H^m), \textrm{SFA}(H^d) \\
    A_ijk &= A^h_i * A^m_j * A^d_k\\
    V &:= V * A
    \end{aligned}
\end{equation*}

\section{Experiment}

\begin{table}
    \centering
    \small
    \scalebox{0.9}{
    \begin{tabular}{clccccc}
    \toprule
    \textbf{Dataset} & \textbf{Model} & \bf Hold. & \bf Tar. & \bf Exp. & \bf Avg. & \bf Targeted\\
    \midrule
    \multirow{4}{*}{\textbf{MultiB}$_\textrm{EU}$} & RACL-B & - & 59.9 & 72.6 & - & 56.8 \\
     & BiAF &  60.4 & 64.0 & 73.9 & 66.1 & 57.8 \\
     & SFA & 65.1 & 68.8 &74.4 & 69.4 & \bf 59.6 \\
     & SFA2o & \bf 65.8 & \bf 71.0 & \bf 76.7 & \bf 71.2 & \bf 59.6 \\
    \midrule
    \multirow{4}{*}{\textbf{MultiB}$_\textrm{CA}$} & RACL-B & - & 67.5 & 70.3 & - & 52.4\\
     & BiAF & 43.0 & 72.5 & 71.1 & 62.2 & 55.0 \\
     & SFA & \bf 52.8 & \bf 74.8 & \bf 73.3 & \bf 67.0 & \bf 62.3 \\
     & SFA2o & 46.2 & 74.2 & 71.0 & 63.8 & 60.9 \\
    \midrule
    \multirow{4}{*}{\textbf{NoReC}$_\textrm{Fine}$} & RACL-B & - & 47.2 & 56.3 & - & 30.3 \\
     & BiAF & 60.4 & 54.8 & 55.5 & 56.9 & \bf 31.9 \\
     & SFA & 62.5 & \bf 59.1 & \bf 58.5 & \bf 60.0 & 31.7 \\
     & SFA2o & \bf 63.6 & 55.3 & 56.1 & 58.3 & 31.5 \\
    \midrule
    \multirow{4}{*}{\textbf{MPQA}} & RACL-B & - & 20.0 & 31.2 & - & 17.8 \\
     & BiAF & 43.8 & 51.0 & 48.1 & 47.6 & 33.5 \\
     & SFA & 44.6 & \bf 51.7 & \bf 49.2 & 48.5 & \bf 34.0 \\
     & SFA2o & \bf 47.9 & 50.7 & 47.8 & \bf 48.8 & 33.7 \\
    \midrule
    \multirow{4}{*}{\textbf{DS}$_\textrm{Unis}$} & RACL-B & - & 44.6 & 38.2 & - & 27.3 \\
     & BiAF & 37.4 & 42.1 & \bf 45.5 & 41.7 & 29.6 \\
     & SFA & 40.0 & 41.9 & 42.7 & 41.5 & 28.6 \\
     & SFA2o & \bf 50.0 & \bf 44.8 & 43.7 & \bf 46.2 & \bf 30.7 \\
    \bottomrule
    \end{tabular}
    }
    \caption{Comparison of model performance on structured sentiment analysis.}
    \label{tab:res_ssa}
\end{table}

\subsection{Dataset and Configuration}

For comparison with previous state-of-the-art model \cite{DBLP:conf/acl/BarnesKOOV20}, we conduct our experiments on datasets of multiple languages, including hotel reviews \textbf{MultiB}$_\textrm{EU}$, \textbf{MultiB}$_\textrm{CA}$ \cite{barnes-etal-2018-multibooked} in Basque and Catalan, professional reviews \textbf{NoReC}$_\textrm{Fine}$ \cite{DBLP:conf/lrec/OvrelidMBV20} in Norwegian, news wire text \textbf{MPQA} \cite{DBLP:journals/lre/WiebeWC05} in English and reviews of online universities and e-commerce \textbf{DS}$_\textrm{Unis}$ \cite{toprak-etal-2010-sentence} in English.

Hidden size for our layers is $100$ for words and features, except $50$ for characters, $400$ for either direction of BiLSTMs, $600$ for edge and label representations. BiLSTMs is of $3$ layers and SFA uses $4$ attention heads and kernel size of $3$ for mean pooling. We use GloVe \cite{DBLP:conf/emnlp/PenningtonSM14} as word embedding and BERT-base-multilingual-cased \cite{DBLP:conf/naacl/DevlinCLT19} is used as pre-trained language model for all datasets for fair comparison. Our parser is trained using cross entropy loss for both edges and labels. Dropout rate \cite{DBLP:journals/jmlr/SrivastavaHKSS14} is set to $0.33$, and our optimizer is Adam \cite{DBLP:journals/corr/KingmaB14} of which initial learning rate is $10^{-3}$ with $5000$ decay steps. 

\subsection{Main Result}

We only show comparison with two state-of-the-art models RACL-BERT (RACL-B) \cite{chen-qian-2020-relation} and Biaffine (BiAF) \cite{barnes-etal-2018-multibooked} in this section as these two models outperform other models on all datasets. 


\begin{table}
    \centering
    \small
    \scalebox{0.9}{
    \begin{tabular}{clcc}
    \toprule
    \textbf{Dataset} & \textbf{Model} & \bf Parsing Graph & \bf Sentiment Graph\\
    \midrule
    \multirow{3}{*}{\textbf{MultiB}$_\textrm{EU}$} & BiAF & 60.0 & 54.7  \\
     & SFA & 63.0 ($\uparrow$ 3.0) & 58.9 ($\uparrow$ 4.2) \\
     & SFA2o & \bf 66.1 ($\uparrow$ 6.1) & \bf 62.7 ($\uparrow$ 8.0) \\
    \midrule
    \multirow{3}{*}{\textbf{MultiB}$_\textrm{CA}$} & BiAF & 62.1 & 56.8  \\
     & SFA & 62.2 ($\uparrow$ 0.1) & 57.0 ($\uparrow$ 0.2) \\
     & SFA2o & \bf 64.5 ($\uparrow$ 2.3) & \bf 59.3 ($\uparrow$ 2.5) \\
    \midrule
    \multirow{3}{*}{\textbf{NoReC}$_\textrm{Fine}$} & BiAF & 37.7 & 31.2  \\
     & SFA & 40.0 ($\uparrow$ 2.3) & \bf 32.7 ($\uparrow$ 1.5) \\
     & SFA2o & \bf 40.4 ($\uparrow$ 2.7) & 31.9 ($\uparrow$ 0.7) \\
    \midrule
    \multirow{3}{*}{\textbf{MPQA}} & BiAF & 36.9 & 17.4  \\
     & SFA & 37.6 ($\uparrow$ 0.7) & \bf 20.4 ($\uparrow$ 3.0) \\
     & SFA2o & \bf 38.6 ($\uparrow$ 1.7) & 19.1 ($\uparrow$ 1.7) \\
    \midrule
    \multirow{3}{*}{\textbf{DS}$_\textrm{Unis}$} & BiAF & 33.9 & 26.5  \\
     & SFA & 34.9 ($\uparrow$ 1.0) & \bf 27.7 ($\uparrow$ 1.2)  \\
     & SFA2o & \bf 35.0 ($\uparrow$ 1.1) & 27.4 ($\uparrow$ 0.9)  \\
    \bottomrule
    \end{tabular}
    }
    \caption{Comparison of performance on relation predicting on parsing graph and sentiment graph between our SFA and Biaffine.}
    \label{tab:res_par}
\end{table}

Our main experiment results are presented in Table~\ref{tab:res_ssa} and~\ref{tab:res_par}, we follow metrics in \cite{barnes-etal-2018-multibooked} to use F1 scores to evaluate span and relation extraction. Table~\ref{tab:res_ssa} shows SFA and SFA2o boost parser performance evaluated by nearly all metrics on all datasets. For exceptions, SFA still keeps comparable performance. There are also extremely significant improvements such as Holder F1 on \textbf{DS}$_\textrm{Unis}$ ($\uparrow$ 12.6), Holder F1 on \textbf{MultiB}$_\textrm{CA}$ ($\uparrow$ 9.8) and Target F1 on \textbf{MultiB}$_\textrm{EU}$ ($\uparrow$ 7.0). Also, the application of SFA and SFA2o leads to a 3.8 F1 improvement on average for the holder, target, and expression, which verifies SFA's efficiency for span extraction. Average improvement ($\uparrow$ 2.1) on targeted F1 (Target + Polarity) also convinces the capacity of SFA to not only extract target spans but capture polarity of sentiment towards them as well. Second-order SFA leads to further performance improvement on some datasets (\textbf{MultiB}$_\textrm{EU}$, \textbf{DS}$_\textrm{Unis}$), but not always. This may be attributed to the simplicity of structured sentiment analysis for higher-level models to work well.

Table~\ref{tab:res_par} shows the evaluation on model performance based on parsing metrics. We calculated metrics only on labeled graphs because of the structural nature of structured sentiment analysis. SFA mechanism improves the parser performance on all datasets, for either parsing graph or sentiment graph. Extreme improvement also exists in \textbf{MultiB}$_\textrm{EU}$, on which SFA boosts the parser by a 6.1 F1 score on the parsing graph and an 8.0 F1 score on the sentiment graph. As SFA is initially designed for better parsing, we can safely conclude that SFA actually captures the nature of parsing graphs transformed from structured sentiment analysis. For parsing, second-order SFA generally leads to further improvement on most datasets as the higher-order mechanism is initially introduced to strengthen parsing performance.

\subsection{Ablation Study}

\begin{table}
    \centering
    \small
    \scalebox{0.9}{
    \begin{tabular}{lcc}
    \toprule
    {\bf Model} & \bf Parsing Graph & \bf Sentiment Graph\\
    \midrule
     SFA2o &  \bf 66.1 & \bf 62.7  \\
     \quad -2o & 65.0 & 60.7  \\
     \quad -1o & 63.3 & 59.0  \\
     SFA &  63.0 & 58.9  \\
    \bottomrule
    \end{tabular}
    }
    \caption{Ablation study for contribution of SFA mechanism on first-order (1o) and second-order (2o) parser.}
    \label{tab:res_abl}
\end{table}

Table~\ref{tab:res_abl} shows our results from the ablation experiment to verify the contribution from the SFA mechanism for first-order and second-order parsing. We experiment on \textbf{MultiB}$_\textrm{EU}$ by directly using scores from first and second-order scorers for prediction without the multiplication with SFA scores. Notice that only the SFA mechanism is removed and the high-order parsing procedure is still retained.

By removing SFA, the performance of the parser drops sharply, which reflects the contribution of SFA to model improvement in main experiments. Thus, SFA contributes to both first-order and second-order parsing. Moreover, removing first-order SFA hinders the performance more than removing second-order SFA. The latter even results in performance close to the SFA parser without the second-order mechanism. This further indicates SFA on first-order is much more critical for improvement than SFA on second-order.

\section{Conclusion}

In this paper, we further develop the unified parsing system for structured sentiment analysis. We discover the sparsity of edge distribution in the parsing graph formulated from structured sentiment analysis. We take advantage of this property and propose a novel sparse fuzzy attention that benefits both graph parsing and sentiment graph construction. Sparse Fuzzy Attention boosts current state-of-the-art parser performance significantly. Moreover, the combination with high order parsing shows SFA can also benefit high order relations scoring for parsing. Our SFA method can also be applied to other NLP tasks to address the edge sparsity issue.

\bibliography{anthology}
\bibliographystyle{acl_natbib}

\clearpage

\appendix

\section{Related Work}
\label{app:rel}

Structured sentiment analysis is a unification of multiple span and relation extraction tasks, while also including tuple labeling. Early works try to formulate subtasks of structured sentiment analysis into independent span extraction or relation extraction \citep{choi-etal-2006-joint,yang-cardie-2012-extracting,katiyar-cardie-2016-investigating,DBLP:journals/is/ZhangWF19}, ignoring polarity labeling. Systems like IMN \cite{he-etal-2019-interactive} and RACL \cite{chen-qian-2020-relation} have been developed towards task unification but still fail to reach a real integral analysis.

Thus, \cite{DBLP:conf/acl/BarnesKOOV20} propose a parsing-based procedure which implements a real unified structured sentiment analysis, where spans, relations and labels are all cast into dependencies between components in context. Our attention mechanism refines this procedure by adapting the parser for rather denser real dependency graphs to sparser and more continuous formulated dependency graphs from structured sentiment analysis.

Current mainstream parser, Biaffine \cite{DBLP:conf/iclr/DozatM17,DBLP:conf/acl/DozatM18} concentrates on every pair of heads and dependents for dependency predicting. However, for structured sentiment analysis, only parts of the context is critic for graph construction. This motivates us to create an attention mechanism which selects a series of prior heads and dependents for parsing and gives the birth to our SFA mechanism.

High order parsing \cite{DBLP:conf/acl/WangHT19,DBLP:conf/ijcnlp/WangT20,DBLP:conf/acl/ZhangLZ20,DBLP:conf/emnlp/LiZWP20} is a popular mechanism for strengthening parser by predicting high order relations. Thus we adapt this mechanism to structured sentiment analysis as parsing and further develop it with SFA.

\end{document}